\documentclass[conference]{IEEEtran}
\IEEEoverridecommandlockouts
\usepackage{cite}
\usepackage{amsmath,amssymb,amsfonts}
\usepackage{algorithmic}
\usepackage{graphicx,caption}
\usepackage{textcomp}
\usepackage{xcolor}
\usepackage{listings}
\usepackage{tabularx,booktabs}
\usepackage[list=true]{subcaption}

\newcolumntype{C}{>{\centering\arraybackslash}X} 

\def\BibTeX{{\rm B\kern-.05em{\sc i\kern-.025em b}\kern-.08em
    T\kern-.1667em\lower.7ex\hbox{E}\kern-.125emX}}

\begin{document}

\title{Learning Optimal Fair Scoring Systems for Multi-Class Classification}
\author{\IEEEauthorblockN{Julien Rouzot}
\IEEEauthorblockA{\textit{LAAS-CNRS,} \\ \textit{Université de Toulouse, CNRS} \\
Toulouse, France \\
jrouzot@laas.fr}
\and
\IEEEauthorblockN{Julien Ferry}
\IEEEauthorblockA{\textit{LAAS-CNRS,} \\ \textit{Université de Toulouse, CNRS} \\
Toulouse, France \\
jferry@laas.fr}
\and
\IEEEauthorblockN{Marie-José Huguet}
\IEEEauthorblockA{\textit{LAAS-CNRS, Université de Toulouse,}\\ \textit{CNRS, INSA} \\
Toulouse, France \\
huguet@laas.fr}
}


\def\dataset{\mathcal{E}}
\def\classifier{\mathcal{C}}
\def\featuresset{\mathcal{D}}
\def\labelsset{\mathcal{K}}
\def\sensiblelabelsset{\mathcal{K}_s}
\def\protectset{\mathcal{A}}
\def\nsamples{N}
\def\nfeatures{D}
\def\nlabels{K}
\def\sensibleindex{i_{s}}

\def\asampleind#1{e_{#1}}
\def\afeatind#1{x_{#1}}
\def\alabelbin#1{y_{#1}}
\def\alabelmulti#1{y_{#1}}
\def\apredbin#1{\hat{y_{#1}}}
\def\apredmulti#1{\hat{y_{#1}}}


\def\mjo#1{\textcolor{cyan!80}{#1}}
\def\changer#1{\textcolor{red}{[#1]}}

\def\mo#1{\textcolor{red}{#1}}
\def\ulr#1{\textcolor{violet}{#1}}
\def\seb#1{\textcolor{blue}{#1}}
\def\jul#1{\textcolor{orange}{#1}}
\def\jr#1{\textcolor{brown}{#1}}

\lstdefinelanguage{RuleListsLanguage}{
  keywords={if, then, else},
  keywordstyle=\color{blue}\bfseries,
  ndkeywords={},
  ndkeywordstyle=\color{darkgray}\bfseries,
  identifierstyle=\color{black},
  sensitive=false,
  comment=[l]{//},
  morecomment=[s]{/*}{*/},
  commentstyle=\color{purple}\ttfamily,
  stringstyle=\color{red}\ttfamily,
  morestring=[b]',
  morestring=[b]"
}

\definecolor{RuleListsLanguageBackgroundColor}{rgb}{0.95, 0.95, 0.95}

\maketitle

\begin{abstract}
Machine Learning models are increasingly used for decision making, in particular in high-stakes applications such as credit scoring, medicine or recidivism prediction.
However, there are growing concerns about these models with respect to their lack of interpretability and the undesirable biases they can generate or reproduce. 
While the concepts of interpretability and fairness have been extensively studied by the scientific community in recent years, few works have tackled the general multi-class classification problem under fairness constraints, and none of them proposes to generate fair and interpretable models for multi-class classification. 
In this paper, we use Mixed-Integer Linear Programming (MILP) techniques to produce inherently interpretable scoring systems under sparsity and fairness constraints, for the general multi-class classification setup. 
Our work generalizes the SLIM \emph{(Supersparse Linear Integer Models)} framework that was proposed by Rudin and Ustun to learn optimal scoring systems for binary classification.
The use of MILP techniques allows for an easy integration of diverse operational constraints (such as, but not restricted to, fairness or sparsity), but also for the building of certifiably optimal models (or sub-optimal models with bounded optimality gap).


\end{abstract}

\begin{IEEEkeywords}
Machine Learning, Fairness, Interpretability, Mixed-Integer Linear Programming, Scoring System, Multi-class Classification
\end{IEEEkeywords}

\section{Introduction}
Machine Learning (ML) models are 
increasingly used for decision making, 
in particular in high-stakes applications such as credit scoring~\cite{dastile2020statistical}, medicine~\cite{erickson2017machine} or recidivism prediction~\cite{rudin2018optimized}. 
In the context of supervised learning, models such as Deep Neural Networks are used commonly for their capacity to accurately analyse and capture complex correlations.
However, such tools have been criticized in the literature for their lack of interpretability and the indesirable biases they can reproduce or introduce~\cite{angwin2016machine,DBLP:journals/natmi/Rudin19}. 

Despite their capabilities to handle large amounts of data, most \textit{black-box} (\emph{i.e.}, non-interpretable) models should be avoided in high-stakes decision settings~\cite{DBLP:journals/natmi/Rudin19}. 
In a nutshell, the interpretability of a ML model can be defined as ``the ability to explain or to present in understandable terms to a human''~\cite{doshi2017towards}. 
This definition is rather abstract, and the quality of any explanation mechanism depends on the task at hand, the context and on the type of user receiving the explanation.

From a technical perspective, two main approaches have emerged in the literature to enhance the comprehensibilty of ML models~\cite{DBLP:journals/cacm/Lipton18}. 
On the one hand, \emph{post-hoc explanations}~\cite{DBLP:journals/csur/GuidottiMRTGP19} can be used \emph{a posteriori} to explain (globally or locally) the decisions of a black-box model. 
While such explanations can be useful in non-sensitive settings, they usually come without trustworthiness guarantees, in particular with respect to how they really reflect the underlying reasoning of the model, and may be manipulated~\cite{DBLP:conf/icml/AivodjiAFGHT19,DBLP:journals/natmi/Rudin19}.
On the other hand, \emph{transparent-box design} methods build models that are interpretable by nature, such as rule-based models, tree-based models or scoring systems~\cite{DBLP:journals/sigkdd/Freitas13}. 
To ensure interpretability, \emph{sparsity} is often encouraged to keep the models' sizes reasonable. 
In addition, because of their apparent simplicity (which is necessary to keep them understandable), \emph{interpretable-by-design} models are often criticized in the literature for their presumed lack of performances (\emph{e.g.}, in terms of accuracy). 
However, it has been demonstrated that they often perform as well as black-box models for a wide range of real-world applications~\cite{DBLP:journals/natmi/Rudin19}.

The problem of algorithmic
bias, in which a trained ML model uses irrelevant or discriminatory features for decision making, has been identified and widely studied~\cite{barocas-hardt-narayanan}.
To tackle it, different families of fairness notions have been proposed, namely individual fairness, causal fairness and statistical fairness \cite{verma2018fairness}. 
In this paper, we consider statistical fairness notions, which have been extensively studied in the last decade. 

Many works have been proposed in recent years, attempting to define interpretability notions~\cite{DBLP:journals/cacm/Lipton18,doshi2017towards}, fairness metrics~\cite{verma2018fairness,barocas-hardt-narayanan} and developing new frameworks that address these important challenges \cite{10.1145/3411764.3445261}.
However, most of these works have focused on binary classification and few solutions have been proposed to handle the more general multi-class setup~\cite{caton2020fairness,pmlr-v54-zafar17a}.
Furthermore, to the best of our knowledge, none of the works addressing fairness in multi-class classification also considered interpretability.
In this paper, we address this issue by proposing a framework to generate fair and interpretable models for multi-class classification. 
Our approach 
produces certifiably optimal models, which is crucial as lack of optimality can have societal implications~\cite{DBLP:journals/jmlr/AngelinoLASR17}.

More precisely, our contributions are as follows:
\begin{enumerate}
    \item We review and summarize in a unified notation common multi-class fairness metrics proposed in the literature, and build on them to formulate our own metrics, introducing the flexible notion of \emph{sensitive labels}. 
    \item We design and implement a new framework based on Mixed Integer Linear Programming to generate optimal sparse scoring systems for multi-class classification.
    \item We integrate fairness constraints in this framework to generate optimal sparse and fair scoring systems for multi-class classification. 
    The resulting method, named FAIRScoringSystems, is available online\footnote{https://gitlab.laas.fr/roc/julien-rouzot/fairscoringsystemsv0}.
    \item We empirically evaluate the effectiveness of the proposed approach to learn interpretable sparse models achieving good trade-offs between accuracy and fairness in multi-class classification problems.
\end{enumerate}

This paper is organized as follows. 
First, in Section~\ref{section:background}, we provide the necessary background on multi-class classification and fairness in ML. 
Then, in Section~\ref{section:related_work}, we present the SLIM framework~\cite{DBLP:journals/ml/UstunR16,rudin2018optimized} that generates optimal scoring systems for binary classification and provide a summary of the multi-class fairness metrics proposed in the literature in a unified notation.
Afterwards, in Section~\ref{section:contributions}, we detail our contributions for fair scoring systems for multi-class classification.
Finally, we present and analyse our experimental results in Section~\ref{section:experiments} before concluding in Section~\ref{section:conclusion}.
\section{Background}\label{section:background}

In this section, we first introduce the notations that are used throughout the paper, and describe the ML classification task. 
Then, we present the relevant performance metrics.
Finally, we define the notion of statistical fairness for binary classification.


\subsection{Classification}

We use the following notations throughout this paper. 
Let $\dataset$ be a dataset containing $\nsamples$ samples: $\dataset=\{\asampleind{i, i \in \{1\ldots\nsamples\}}\}$.
Each sample $\asampleind{i} \in \dataset$ is defined by the values of $\nfeatures$ binary features $(\afeatind{i,1}, \ldots, \afeatind{i,\nfeatures})$ and one label $\alabelmulti{i} \in \labelsset$, with $\lvert \labelsset \rvert = \nlabels$. If $\alabelmulti{i} = k$, we say that $\asampleind{i}$ belongs to class $k$.
Finally, let $\apredmulti{i}$ denotes the label predicted by a trained ML model for sample $\asampleind{i}$.

In ML, a classification task refers to a predictive problem in which, given some input features $(\afeatind{i,1}, \ldots, \afeatind{i,\nfeatures})$, a \emph{model} aims to predict the label associated to this input. 
Supervised learning methods process a labelled dataset and exploit the correlations learnt from the data to produce a model.
Given a sufficient amount of training data, 
the objective is to accurately predict (through $\apredmulti{i}$) the label $\alabelmulti{i}$ of a new sample~\cite{saravanan2018state}. 
A common classification problem is binary classification, where $\nlabels = 2$ (\emph{e.g.,} is an email ``spam'' or ``not spam''). 
In this paper, we will focus on the more general multi-class classification, in which the output of the classification algorithm can take any value in a set $\labelsset$ of possible classes. 

Two main approaches can be used to aggregate binary classification models to tackle multi-class classification tasks~\cite{aly2005survey}.
The \textit{one-vs-all} strategy aims to fit one binary classifier per class, distinguishing the class' examples from those of the other classes. 
The \textit{all-vs-all} strategy (also called \emph{one-vs-one}) constructs one classifier for each pair of classes.
The later often leads to a lack of interpretability as the \textit{all-vs-all} strategy produces much more classifiers, so we consider the \textit{one-vs-all} approach in our framework. 

\begin{table*}[ht]
 \caption{Standard fairness metrics for binary classification.}
\label{tab:binary-fairness-metrics}
\begin{tabularx}{\textwidth}{@{}l*{2}{C}c@{}}
\toprule
Metric & Mathematical expression\\ 
\midrule
Statistical Parity~\cite{10.1145/2090236.2090255}&$P(\apredbin{i}=1 \:|\: \asampleind{i} \in \protectset) = P(\apredbin{i}=1 \:|\: \asampleind{i} \notin \protectset)$\\
        
Overall Misclassification Rate~\cite{ye2020unbiased}&$P(\apredbin{i} \neq y_{i} \:|\: \asampleind{i} \in \protectset) = P(\apredbin{i} \neq y_{i} \:|\: \asampleind{i} \notin \protectset)$\\ 

Predictive Equality~\cite{chouldechova2017fair}&$P(\apredbin{i} = 1 \:|\: \asampleind{i} \in \protectset, \alabelmulti{i} = 0) = P(\apredbin{i} = 1 \:|\: \asampleind{i} \not\in \protectset, \alabelmulti{i} = 0)$\\
        
Equal Opportunity~\cite{hardt2016equality}&$P(\apredbin{i} = 1 \:|\: \asampleind{i} \in \protectset, \alabelmulti{i} = 1) = P(\apredbin{i} = 1 \:|\: \asampleind{i} \not\in \protectset, \alabelmulti{i} = 1)$\\ 
        
Equalized Odds~\cite{hardt2016equality}&$P(\apredbin{i} = 1 \:|\: \asampleind{i} \in \protectset, \alabelmulti{i}) = P(\apredbin{i} = 1 \:|\: \asampleind{i} \not\in \protectset, \alabelmulti{i})$\\
\bottomrule
\end{tabularx}
\end{table*}


\subsection{Performance Metrics} 
To evaluate the performances of a multi-class classification model, its \textit{confusion matrix} is commonly used. 
In this $\nlabels \times \nlabels$ matrix, each row corresponds to an actual label and each column represents a predicted label.
Each cell $(k_a, k_b)$ contains the number of samples belonging to the class $k_a$ in the dataset and predicted in the class $k_b$ by the model. Hence, the diagonal of this matrix represents the samples that have been correctly classified. 
Based on these values, a popular metric for assessing a model's performance is its predictive \emph{accuracy}~\cite{Grandini2020MetricsFM}, 
which evaluates the proportion of correctly classified samples: 

\begin{equation}
    \text{Accuracy} = \frac{\sum_{k=1}^{K}, TP_k}{N}\label{metric:accuracy}
\end{equation}

in which $TP_k$ represents the number of \emph{True Positive} examples for a given class $k$: $TP_k = \lvert \{ \asampleind{i} \mid  \alabelmulti{i} = k \land \apredmulti{i} = k\} \rvert$.

However, when dealing with imbalanced datasets, the \emph{balanced accuracy} metric is often preferred.
In a nutshell, it computes the model's average accuracy over the different classes:

\begin{equation}
    \text{Balanced Accuracy} = \frac{\sum_{k=1}^{K}, \frac{TP_k}{N_k}}{K}\label{metric:balanced_accuracy}
\end{equation}

in which $N_k$ is the number of samples belonging to class $k$: $N_k = \lvert \{ \asampleind{i} \mid  \alabelmulti{i} = k\} \rvert$.


\subsection{Fairness} 

Statistical fairness (also called group fairness) notions aim to correct a prediction bias between different subsets of a dataset called \textit{protected groups}. 
These protected groups usually differ by the value of one or several protected feature(s) (\emph{e.g.}, age, gender or ethnicity)~\cite{NEURIPS2021_32e54441}. 
In this paper, we let $\protectset \subset \dataset$ be a protected group, and the purpose of fairness notions is to ensure that the learnt classifier behaves similarly between individuals from protected group $\protectset$ and from the rest of the population ($\dataset \setminus \protectset$).
To realize this, fairness metrics are used to measure the unfairness of a classifier based on a certain statistical criterion.
While different fairness metrics have been introduced, some of them are incompatible~\cite{verma2018fairness}, meaning that they cannot be optimized altogether. 

We first introduce widely used fairness metrics in the binary classification setup ($\labelsset = \{0, 1\}$) in Table~\ref{tab:binary-fairness-metrics}. 
In a nutshell, statistical parity ensures that individuals from the two groups have the same probabilities to be positively predicted.
Overall misclassification rate equalizes the probabilities of being incorrectly classified.
Predictive equality equalizes the False Positive Rates (FPR), while equal opportunity equalizes the True Positive Rates (TPR).
Finally, equalized odds can be interpreted as the conditional independence between the prediction of the classifier and the protected feature given the ground truth (and is, in this setup, the conjunction of predictive equality and equal opportunity).

As exact fairness is often too restrictive, a common relaxation consists in quantifying the difference of some statistical measure among the different protected groups.
The model at hand is then coined as \emph{fair} if its measured fairness violation (unfairness)  
is lower than an \emph{unfairness tolerance} $\epsilon$. 
In practice, probabilities are estimated using empirical rates.

%
\section{Related Work}\label{section:related_work}
In this section, we first introduce the SLIM framework~\cite{DBLP:journals/ml/UstunR16,rudin2018optimized} that was proposed to learn optimal scoring systems for binary classification. 
Then, we review existing works on fairness in multi-class classification before summarizing the existing metrics in a unified notation. 

\subsection{Learning Optimal Scoring Systems for Binary Classification with SLIM}\label{subsec:related_work_slim}

Rudin and Ustun~\cite{DBLP:journals/ml/UstunR16,rudin2018optimized} have developed the \emph{Supersparse Linear Integer Model} (SLIM) framework, which produces optimal (in terms of accuracy and sparsity) scoring systems for binary classification. 
Scoring systems are considered interpretable-by-design models and are deployed in 
real-world applications such as medicine~\cite{DBLP:journals/ml/UstunR16} or crime prediction~\cite{andrade2009handbook}.

\begin{figure}[htb]
    \centering
    \includegraphics[width=\linewidth]{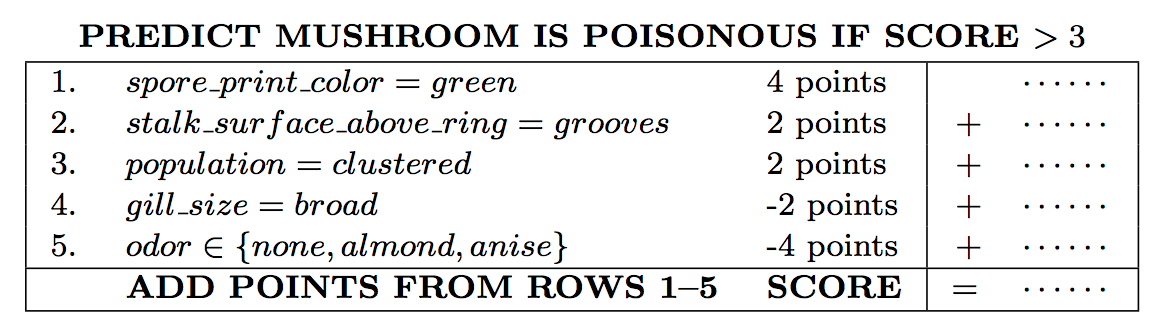}
    \caption{Example scoring system for mushroom toxicity detection generated by SLIM~\cite{DBLP:journals/ml/UstunR16}.}
    \label{fig:scoring_system_binary}.
\end{figure}

As illustrated in Figure~\ref{fig:scoring_system_binary}, a scoring system can be represented as a table in which each row associates a Boolean condition over the datasets' features to a number of points (which can be negative).
If a new sample satisfies the given condition, the associated number of points is added to the score of the sample. 
If the final score is above a given threshold, the sample is predicted to belong to the positive class, otherwise the model predicts the negative one. 

SLIM is based on a Mixed Integer Linear Program (MILP), which can be solved using any off-the-shelf solver. 
While fairness requirements are not originally part of the SLIM framework, it is possible to incorporate (linearized) fairness constraints (as well as other operational constraints~\cite{DBLP:journals/ml/UstunR16}), thanks to the declarative nature of the method.

\subsection{Statistical Fairness in Multi-Class Classification}\label{sec:fairness-multi-class}

While most of the fairness literature focuses on the binary classification setup, some recent works have also studied fairness requirements for multi-class classification. Table~\ref{tab:related-work-multiclass-fairness-metrics} summarizes the proposed fairness metrics using a unified notation. 
While most of the fairness metrics for multi-class classification are straightforward extensions of the binary metrics defined in Table~\ref{tab:binary-fairness-metrics}, equalized odds can be interpreted in different ways in the multi-class setup. 
In particular, multi-class equalized odds (also called term-by-term equality of odds~\cite{putzel2022blackbox}) aims at equalizing the probabilities of being predicted $k\in\labelsset$ given true label $k'\in\labelsset$, for all pairs $(k,k') \in \labelsset^2$, which requires $K^2$ fairness constraints. 
In contrast, classwise equality of odds~\cite{putzel2022blackbox} considers all false positive classifications altogether, which only requires $2 \cdot K$ constraints.

Similarly to the binary classification case, fairness-enhancing techniques for multi-class classification can be categorized into three different categories, depending on which stage of the ML pipeline they intervene~\cite{bellamy2019ai}:

\begin{itemize}

    \item \textit{Pre-processing} methods modify the input dataset to remove undesired correlations with the sensitive feature(s) and/or to satisfy some statistical criteria~\cite{romano2020achieving,https://doi.org/10.48550/arxiv.2205.15860}.
    
    \item \textit{Post-processing} methods modify the output of the learning algorithm, such as a trained classifier, to satisfy some fairness criteria~\cite{putzel2022blackbox,yang2020fairness,https://doi.org/10.48550/arxiv.2206.07801}.
    
    \item Our framework belongs to the last category, which corresponds to \textit{in-processing} (also called \emph{algorithmic modification}) approaches. 
    These methods tackle the unfair biases directly during the learning process \cite{ye2020unbiased,yang2020fairness,hossain2020designing,denis:hal-03355938}, which generally leads to the best fairness/utility tradeoffs~\cite{barocas-hardt-narayanan}.
\end{itemize}

To the best of our knowledge, fairness and interpretability have not been tackled altogether in any work for multi-class classification.

\begin{table*}[h!]
\caption{Standard fairness metrics for multi-class classification.}
\label{tab:related-work-multiclass-fairness-metrics}
\begin{tabularx}{\textwidth}{@{}l*{2}{C}c@{}}
\toprule
Metric & Mathematical expression\\ 
\midrule
Statistical Parity (or Demographic Parity)~\cite{denis:hal-03355938,https://doi.org/10.48550/arxiv.2206.07801,https://doi.org/10.48550/arxiv.2205.15860,yang2020fairness} & $P(\apredbin{i} = k \:|\: \asampleind{i} \in \protectset) = P(\apredbin{i} = k \:|\: \asampleind{i} \not\in \protectset) \quad \forall k \in \labelsset$\\
        
Equal Opportunity (or Overall Accuracy)~\cite{yang2020fairness,https://doi.org/10.48550/arxiv.2206.07801} & $P(\apredbin{i} = k \:|\: \asampleind{i} \in \protectset,  \alabelmulti{i} = k) = P(\apredbin{i} = k \:|\: \asampleind{i} \not\in \protectset,  \alabelmulti{i} = k) \quad \forall k \in \labelsset$\\
   
Overall Misclassification Rate (or Max Loss Discrepancy)~\cite{ye2020unbiased} & $P(\apredbin{i} \neq k \:|\: \asampleind{i} \in \protectset, \alabelmulti{i} = k) = P(\apredbin{i} \neq k \:|\: \asampleind{i} \not\in \protectset, \alabelmulti{i} = k) \quad \forall k \in \labelsset$\\ 

Equalized Odds~\cite{romano2020achieving,https://doi.org/10.48550/arxiv.2206.07801,putzel2022blackbox} & $P(\apredbin{i} = k \:|\: \asampleind{i} \in \protectset, \alabelmulti{i}) = P(\apredbin{i} = k \:|\: \asampleind{i} \not\in \protectset, \alabelmulti{i}) \quad \forall k \in \labelsset$\\

Classwise Equality of Odds~\cite{putzel2022blackbox} & $P(\apredbin{i} = k \:|\: \alabelmulti{i} \neq k, \asampleind{i} \in \protectset) = P(\apredbin{i} = k \:|\: \alabelmulti{i} \neq k, \asampleind{i} \not\in \protectset) \quad \text{AND}$\\ 
~&$P(\apredbin{i} = k \:|\: \alabelmulti{i} = k, \asampleind{i} \in \protectset) = P(\apredbin{i} = k \:|\: \alabelmulti{i} = k, \asampleind{i} \not\in \protectset) \quad \forall k \in \labelsset$\\ 
\bottomrule
\end{tabularx}
\end{table*}
\section{Learning Optimal Fair and Sparse Scoring Systems for Multi-Class Classification}\label{section:contributions}
In this section, we present our novel approach for learning optimal scoring systems for multi-class classification under fairness and sparsity constraints.
First, we describe the considered fairness metrics for multi-class classification, introducing the novel concept of \emph{sensitive labels}.
Then, we extend binary scoring systems (as produced by SLIM) to the multi-class setting using the \emph{one-vs-all} approach.
Finally, we introduce FAIRScoringSystems, a flexible MILP framework to generate fair and interpretable models for multi-class classification.   

\subsection{Proposed Fairness Metrics for Multi-Class Classification}\label{section:contributions_metrics}

Applying binary fairness metrics (as defined in Table~\ref{tab:binary-fairness-metrics}) on each label $k \in \labelsset$ is not always necessary for all use cases and may sometimes lead to poor performances when unnecessary constraints are applied.
We propose a more generic setup in which the set of labels $\labelsset$ is partitioned into a subset of \emph{sensitive labels} $\sensiblelabelsset \subseteq \labelsset$ and a subset of \emph{unsensitive labels} ($\labelsset \setminus \sensiblelabelsset$). 
Intuitively, 
sensitive labels correspond to outcomes yielding great impact on individuals' lives for the decision-making process in which the scoring system is applied.

Afterwards, our proposed fairness metrics for multi-class classification only apply on the sensitive labels. 
More precisely, given a subset of sensitive labels $\sensiblelabelsset$ and a protected group $\protectset$, we apply the binary metrics defined in Table~\ref{tab:binary-fairness-metrics} on each sensitive label $k \in \sensiblelabelsset$.
The resulting metrics are defined in Table~\ref{tab:proposed-multi-fairness-metrics}. 
Note that some standard fairness metrics introduced in Section~\ref{sec:fairness-multi-class} correspond to the special case in which $\sensiblelabelsset = \labelsset$. 
In particular, our multi-class Equalized Odds metric corresponds to the classwise equality of odds metric~\cite{putzel2022blackbox}. 
As Overall Misclassification Rate is equivalent to Equal Opportunity in our setup (balancing the True Positive Rate for class $k$ is equivalent to balancing its False Negative Rate), we do not consider this metric for our framework. 


\begin{table*}[ht]
\caption{Proposed fairness metrics for multi-class classification, using the notion of \emph{sensitive labels}}
\label{tab:proposed-multi-fairness-metrics}
\begin{tabularx}{\textwidth}{@{}l*{2}{C}c@{}}
\toprule
Metric & Mathematical expression\\ 
\midrule
Multi-class Statistical Parity (SP)&$P(\apredbin{i}=k \:|\: \asampleind{i} \in \protectset) = P(\apredbin{i}=k \:|\: \asampleind{i} \notin \protectset) \quad \forall k \in \sensiblelabelsset$\\ 
            
Multi-class Predictive Equality (PE)&$P(\apredbin{i}=k \:|\: \asampleind{i} \in \protectset, \alabelbin{i} \neq k) = P(\apredbin{i}=k \:|\: \asampleind{i} \notin \protectset, \alabelbin{i} \neq k) \quad \forall k \in \sensiblelabelsset$\\

Multi-class Equal Opportunity (EO)&$P(\apredbin{i}=k \:|\: \asampleind{i} \in \protectset, \alabelbin{i}=k) = P(\apredbin{i}=k \:|\: \asampleind{i} \notin \protectset, \alabelbin{i}=k) \quad \forall k \in \sensiblelabelsset$\\ 

Multi-class Equalized Odds (EOD)&Predictive Equality AND Equal Opportunity\\
\bottomrule
\end{tabularx}
\end{table*}


\subsection{Scoring Systems for Multi-Class Classification}
We extend binary scoring systems (as originally produced by SLIM) to multi-class classification using the \emph{one-vs-all} paradigm~\cite{aly2005survey}. 
More precisely, one scoring system is generated for each label $k \in \labelsset$ of the dataset, whose purpose is to detect examples belonging to class $k$.
To classify a new sample $\asampleind{i}$, each scoring system is applied and the class corresponding to the scoring system with the highest score is predicted.

\begin{figure}[h]
    \centering
   \includegraphics[width=0.5\textwidth]{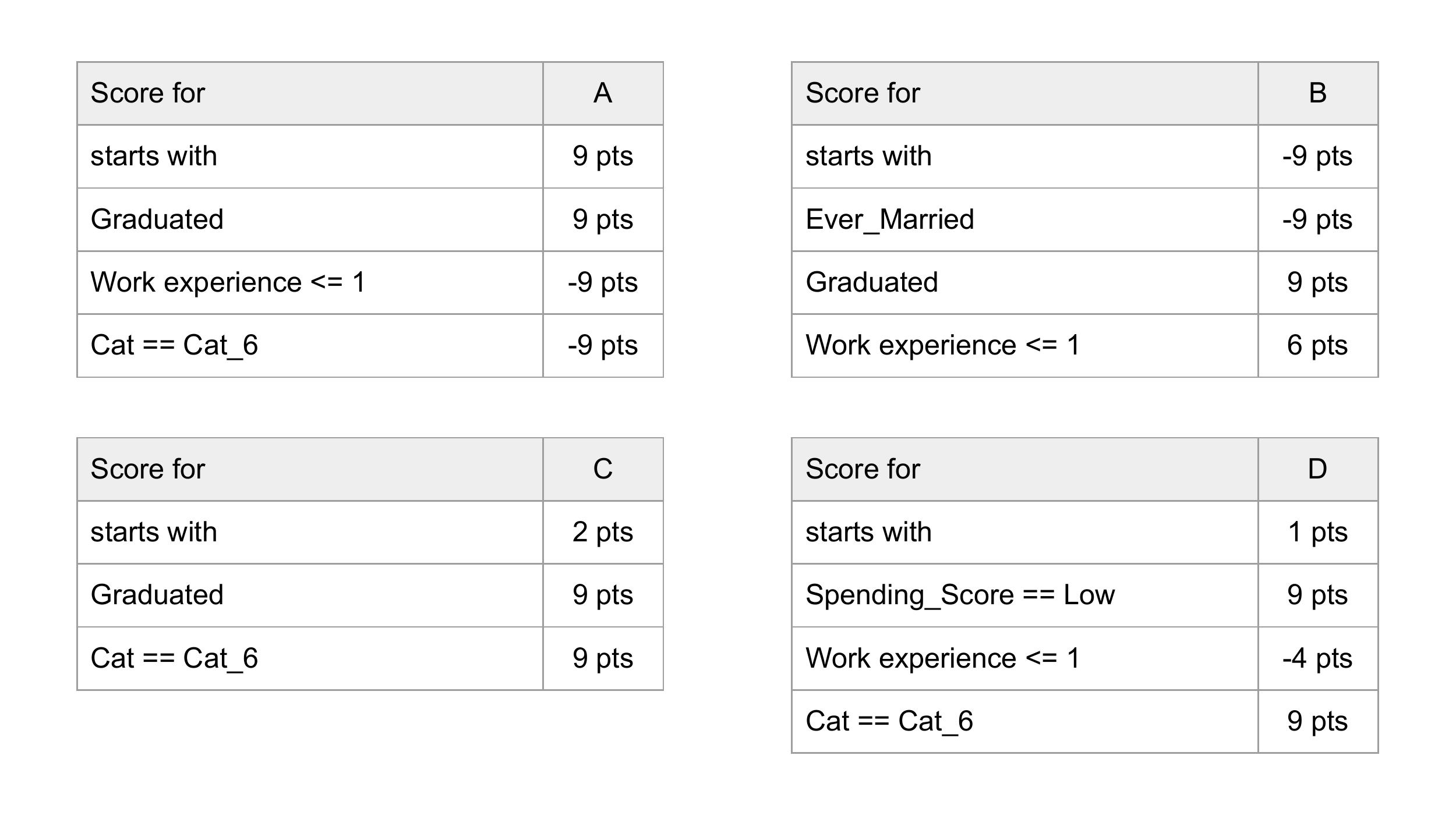}
    \caption{Example multi-class scoring system generated by FAIRScoringSystems for customer segmentation with sparsity (number of lines in each scoring system less or equal to $4$) and fairness (equal opportunity violation less or equal to $0.01$) constraints. 
    This classifier exhibits an accuracy of 0.4495 and an equal opportunity violation of 0.0057 ($\protectset=\{\asampleind{i}\:|\:\afeatind{i,\text{sex=female}} = 1\}$, $\sensiblelabelsset=\{\text{A}\}$).
    ``starts with" can be seen as a virtual attribute, which is True for all examples. It introduces a constant bias inherent to each individual scoring system.}
    \label{fig:multi-class-scoring-systems}
\end{figure}

Figure \ref{fig:multi-class-scoring-systems} provides an example of multi-class scoring system generated for the \textit{customer} dataset. 
This dataset contains $4$ different labels ($\labelsset = \{A, B, C, D\}$) corresponding to customers' categories and one scoring system is generated for each one of them. 
Given a new sample $\asampleind{i}$, each scoring system is applied and the predicted label is the one whose scoring system yields the highest score.

\subsection{FAIRScoringSystems}
Hereafter, we describe FAIRScoringSystems, a Mixed Integer Linear Program (MILP) extending the SLIM framework~\cite{DBLP:journals/ml/UstunR16,rudin2018optimized} (introduced in Section~\ref{subsec:related_work_slim}) to the multi-class classification setup.  
More precisely, FAIRScoringSystems generates optimal multi-class scoring systems, maximizing accuracy~\eqref{fairscoringsystems:accuracy} or balanced-accuracy~\eqref{fairscoringsystems:balanced_accuracy}, given fairness and sparsity constraints. 
Due to the declarative nature of the approach, such constraints can be tuned by the user and additional operational constraints can easily be handled.
The proposed MILP formulation is detailed hereafter.

\subsubsection{Variables} 
\begin{itemize}
    \item 
    $\lambda_{k\in\{1..K\},j\in\{1..\nfeatures\}} \in \mathcal{L}$: matrix $(\nlabels, \nfeatures)$ of integer variables representing the value of each coefficient $j \in \{1..\nfeatures\}$ for each scoring system $k \in \{1..K\}$. 
    These \emph{decision variables}  
    fully describe our multi-class scoring system.
    Each coefficient $\lambda_{k,j}$ belongs to a user-defined set of integers $\mathcal{L}$, lower-bounded (respectively upper-bounded) by $\mathcal{L}_{min}$ (respectively $\mathcal{L}_{max}$).
    \item $z_{i\in\{1..\nsamples\}} \in \{0,1\}$: 
    binary loss variables, each representing the misclassification of a training sample $\asampleind{i\in\{1..\nsamples\}}$. 
    \item $\alpha_{k\in\{1..K\},j\in\{1..\nfeatures\}}  \in \{0,1\}$: matrix $(\nlabels, \nfeatures)$ of binary variables indicating if each coefficient $\lambda_{k,j}$ is non-zero. 
    As null coefficients do not appear in the final scoring system, the sparsity of the model directly depends on the number of non-zero coefficients. 
    \item $pos_{i\in\{1..\nsamples\},k\in\{1..K\}} \in \{0,1\}$: matrix $(\nsamples, \nlabels)$ of binary variables modelling the (one-hot encoded) prediction $\apredbin{i}$ of our classifier for each sample $\asampleind{i\in\{1..\nsamples\}}$ of the training set, as this information is explicitly required to formulate fairness constraints. 
\end{itemize}

\subsubsection{Objective}
The objective of our MILP, given in \eqref{fairscoringsystems:obj}, is to minimize the misclassification error $L$ along with a regularisation term $\gamma_{reg}$  encouraging sparsity.

\begin{align}
    \min \quad L + \gamma_{reg} \label{fairscoringsystems:obj}    
\end{align}

The misclassification error $L$ can either be the empirical loss \eqref{fairscoringsystems:accuracy} or the balanced empirical loss \eqref{fairscoringsystems:balanced_accuracy}. 
This is equivalent to maximizing the accuracy~\eqref{metric:accuracy} or the balanced accuracy~\eqref{metric:balanced_accuracy}.

\begin{align}
    & L = \frac{1}{N} \cdot \sum_{i=1}^{N} z_i\label{fairscoringsystems:accuracy}\\
   \text{or} \quad & L = \sum_{k=1}^K \left( \frac{1}{N_k} \sum_{i | y_i = k} z_i \right)\label{fairscoringsystems:balanced_accuracy}
\end{align}

The regularisation term $\gamma_{reg}$, defined in \eqref{fairscoringsystems:gamma_reg}, is weighted by a constant to ensure that the model will choose a sparser model (as the sum of the non-zero coefficients is minimized) only if it does not affect the loss.
Thus, $\gamma_{reg}$ must be strictly smaller than the additional loss implied by a single misclassification. 
As $\max \sum_{k=1}^{K} \sum_{j=1}^{D} \alpha_{k,j} = D \cdot K$ and as a single misclassification adds $\frac{1}{\min_{k\in\labelsset} (N_k)}$ to the balanced empirical loss (with $\min_{k\in\labelsset} (N_k)$ the number of samples in the most underrepresented class), we have (in the case of the empirical loss, $\min_{k\in\labelsset} (N_k)$ is replaced by $N$):
 
\begin{align}
    & \gamma_{reg} = \frac{1}{\min_{k\in\labelsset} (N_k) \cdot D \cdot K + 1} \cdot \sum_{k=1}^{K} \sum_{j=1}^{D} \alpha_{k,j}\label{fairscoringsystems:gamma_reg}.
\end{align}

\subsubsection{Constraints}

\paragraph*{Classification-related constraints}
\begin{align}
    &-M_i z_i \leq s_{i,k} - s_{i,k'} & i \in 1..N \quad y_i = k\nonumber\\
    &&(k,k') \in (1..K)^2 \quad k \neq k'\label{fairscoringsystems:z}\\
    & s_{i,k} = \sum_{j=1}^{D} \lambda_{k,j} \cdot x_{i,j} - \gamma k & i \in 1..N \quad k \in 1..K\label{fairscoringsystems:score}
\end{align}

\paragraph*{Sparsity-related constraints}
\begin{align}
    & \mathcal{L}_{max} \cdot \alpha_{k,j} \geq \lambda_{k,j} & k \in 1..K \quad j \in 1..D \label{fairscoringsystems:alpha_1}\\
    & \mathcal{L}_{min} \cdot \alpha_{k,j} \leq -\lambda_{k,j} & k \in 1..K \quad j \in 1..D \label{fairscoringsystems:alpha_2}\\
    & \sum_{j=1}^D \alpha_{k, j} \leq \epsilon_s & k \in 1..K \label{fairscoringsystems:sum_alpha}
\end{align}

\paragraph*{Predictions-modeling constraints}
\begin{align}
    &-M_i (1 - pos_{i,k})\leq s_{i,k} - s_{i,k'} \nonumber\\ 
    &\quad\quad\quad\quad\quad\quad\quad i \in 1..N \quad (k,k') \in (1..K)^2 \quad k \neq k'\label{fairscoringsystems:pos}\\
    & \sum_{k=1}^K pos_{i,k} = 1 \quad i \in 1..N \label{fairscoringsystems:sum_pos}
\end{align}

\paragraph*{Statistical parity constraints}
\begin{align}
    &\left|\frac{1}{\lvert \protectset \rvert} \sum_{\asampleind{i} \in \protectset} pos_{i,k} - \frac{1}{\lvert \dataset \backslash \protectset \rvert} \sum_{\asampleind{i} \not\in \protectset} pos_{i,k}\right| \leq \epsilon_{sp} &  k \in \sensiblelabelsset \label{fairscoringsystems:sp}
\end{align}

\paragraph*{Predictive Equality constraints}
\begin{align}
    &\left| \frac{1}{\lvert \protectset \rvert} \sum_{\asampleind{i} \in \protectset, y_i \neq k} pos_{i,k} - \frac{1}{\lvert \dataset \backslash \protectset \rvert} \sum_{\asampleind{i} \not\in \protectset, y_i \neq k} pos_{i,k}\right| \leq \epsilon_{pe} &  k \in \sensiblelabelsset \label{fairscoringsystems:pe}
\end{align}

\paragraph*{Equal opportunity constraints}
\begin{align}
    &\left| \frac{1}{\lvert \protectset \rvert} \sum_{\asampleind{i} \in \protectset, y_i = k} pos_{i,k} - \frac{1}{\lvert \dataset \backslash \protectset \rvert} \sum_{\asampleind{i} \not\in \protectset, y_i = k} pos_{i,k}\right| \leq \epsilon_{eo} &  k \in \sensiblelabelsset \label{fairscoringsystems:eo}
\end{align}

\paragraph*{Equalized odds constraints} 
Equalized odds is formulated as the conjunction of the predictive equality~\eqref{fairscoringsystems:pe} and the equal opportunity~\eqref{fairscoringsystems:eo} constraints, with $\epsilon_{pe}=\epsilon_{eod}$ et  $\epsilon_{eo}=\epsilon_{eod}$.


Constraints~\eqref{fairscoringsystems:z} and~\eqref{fairscoringsystems:score} allow to model the variable $z_i$ representing the classification errors. 
Constraint~\eqref{fairscoringsystems:score} keeps track of the score $s_{i,k}$ of each sample for each label's individual scoring system $k\in\{1..K\}$. 
A Big-M constraint~\eqref{fairscoringsystems:z} allows to set $z_i$ to 1 if the associated sample is misclassified ($s_{i,k} < s_{i,k'} \quad y_i = k$). 
The value of $M_i$ must be greater than $\max (s_{i,k} - s_{i,k'})$. 
A penalty $\gamma k$ is added to apply the lexicographical order in case the scores are the same for two different labels (arbitrary break of ties). 
As the ground truth $y_i = k$ is known, $\hat{y_i} = y_i$ ($\asampleind{i}$ is correctly classified) implies that $s_{i,k} > s_{i,k'} \quad \forall \: k' \neq k$.

To model the variables $\alpha_{k,j}$, we use constraints \eqref{fairscoringsystems:alpha_1} and \eqref{fairscoringsystems:alpha_2} that will set $\alpha_{k,j}$ to 1 if its associated coefficient $\lambda_{k,j} \neq 0$. 
Otherwise, $\alpha_{k,j}$ is set to $0$ as 
$\mathcal{L}_{max} \geq \lambda_{k,j}$ and $\mathcal{L}_{min} \leq -\lambda_{k,j}$. 
The constraint~\eqref{fairscoringsystems:sum_alpha} is applied to each label's scoring system to impose a sparsity constraint by limiting the maximal number of non-zero coefficients. 
The sparsity limit $\epsilon_s$ is a parameter.

The variable $pos_{i,k}$ represents the (one-hot encoded) decision $\apredbin{i}$ of the multi-class scoring system for each sample $\asampleind{i\in\{1..\nsamples\}}$ of the training set (required to express our fairness constraints).
To model this variable, we set $pos_{i,k}$ to $0$ if the associated score is lower than at least one other score~\eqref{fairscoringsystems:pos} and let $pos_{i,k}$ free otherwise.
Constraint~\eqref{fairscoringsystems:sum_pos} ensures that the variable $pos_{i,k}$ associated with the highest score is set to 1.


The fairness constraints are then defined according to the user choice. 
In this paper, we focus on the most popular fairness metrics, but more operational constraints can be easily implemented. 
In line with state-of-the-art methods, our constraints bound the fairness violation (for protected group $\protectset$)
by some unfairness tolerance $\epsilon_{f}$ set by the user, in which $f$ represents the chosen fairness metric. 
In practice, each fairness constraint is implemented as a conjunction of two linear constraints (to get rid of the non-linear absolute value).


%
\section{Experiments}\label{section:experiments}
In this section, we empirically evaluate FAIRScoringSystems using one synthetic and two real-world datasets. 
First, we introduce the experimental setup before evaluating the resulting trade-offs between accuracy (or balanced accuracy) and sparsity.
Finally, we assess the effectiveness of the method to produce interesting trade-offs between accuracy (or balanced accuracy) and fairness under different sparsity constraints. 

\subsection{Setup}
We compare the performances of FAIRScoringSystems with two baseline methods: \texttt{FAIR} (fair baseline) and \texttt{SVM} (accurate baseline). 
More precisely, \texttt{FAIR} corresponds to a constant classifier predicting the majority label. 
It is the trivial model with highest accuracy, exhibiting perfect fairness (unfairness $0$) for all our considered metrics.
In contrast, \texttt{SVM} is a standard one-vs-all multi-class linear kernel SVM (without fairness constraints) using the scikit-learn\footnote{https://scikit-learn.org} implementation. 
The value of the regularization hyper-parameter $C$ is optimized using a standard grid-search technique with a validation set.
While often considered as a black-box model~\cite{DBLP:journals/sigkdd/Freitas13}, \texttt{SVM} offers an intuitive accuracy baseline because it consists in an aggregation of linear models, similarly to our proposed multi-class scoring systems.
All method are evaluated using one synthetic and two real-world datasets: 

\begin{itemize}

    \item \textit{Synthetic dataset} ($\nsamples=800$, $\nfeatures=6$, $\labelsset=\{\text{L1, L2, L3}\}$, $\protectset=\{\asampleind{i}\:|\:\afeatind{i,\text{A1}} = 1\}$, $\sensiblelabelsset=\{\text{L1}\}$): The features are generated with a random distribution between 40\% and 60\% of 1, and 3 labels L1, L2 and L3 are computed from the features (L1 depends on features 1-2-3, L2 on features 3-4-5, L3 on features 5-6-1) with a random noise\footnote{https://gitlab.laas.fr/roc/julien-rouzot/fairscoringsystemsv0}. 
    By construction, this dataset is biased towards labels L1 and L3 for the sensitive feature A1.
    
    \item \textit{wine\footnote{https://www.kaggle.com/datasets/rajyellow46/wine-quality}} ($\nsamples=6497$, $\nfeatures=25$, $\labelsset=\{\text{good, medium, bad}\}$, $\protectset=\{\asampleind{i}\:|\:\afeatind{i,\text{colour=red}} = 1\}$, $\sensiblelabelsset=\{\text{medium}\}$) : This dataset associates chemical characteristics of wines to their quality (good, medium or bad). 
    For the sake of illustration, the considered sensitive feature is the colour of the wine (red or white).
    \item \textit{customer\footnote{https://www.kaggle.com/datasets/vetrirah/customer}} ($\nsamples=6665$, $\nfeatures=29$, $\labelsset=\{\text{A, B, C, D}\}$, $\protectset=\{\asampleind{i}\:|\:\afeatind{i,\text{sex=female}} = 1\}$, $\sensiblelabelsset=\{\text{A, B, C, D}\}$) : This dataset deals with customer segmentation, according to their profile. There are four anonymous customer categories (A, B, C and D), and the sensitive feature is gender (male or female).
\end{itemize}
Following standard procedures, both real-world datasets are binarized and potential inconsistencies are removed as preprocessing. 
As the wine dataset is highly imbalanced (77\% of the samples are labelled ``medium''), we use balanced accuracy as the performance metric for this dataset, while we use accuracy for the other datasets.  

FAIRScoringSystems' MILP is implemented and solved using the IBM ILOG CPLEX 20.1.0.0 solver\footnote{https://www.ibm.com/docs/en/icos/20.1.0} via the \textit{DOcplex}\footnote{http://ibmdecisionoptimization.github.io/docplex-doc/} Python Modeling API and its default configuration. 
For all experiments, we set the integer coefficients' 
range $\mathcal{L} = [-9, 9]$. Restricting coefficients to single-digit numbers enhances interpretability while still allowing sufficient expressivity.

The experiments are conducted on a computing grid over a set of homogeneous nodes using Intel Xeon E5-2695 v4 @ 2.10GHz CPU.
Each experiment runs on a single CPU core for a fixed time limit of one hour and an allocated memory of 20 GB.
Results are averaged using 5-folds cross-validation. 

\subsection{Accuracy/Sparsity Trade-offs}
\begin{figure*}[!htb]
    \centering
    \subcaptionbox{synthetic dataset}{\includegraphics[width=0.31\textwidth]{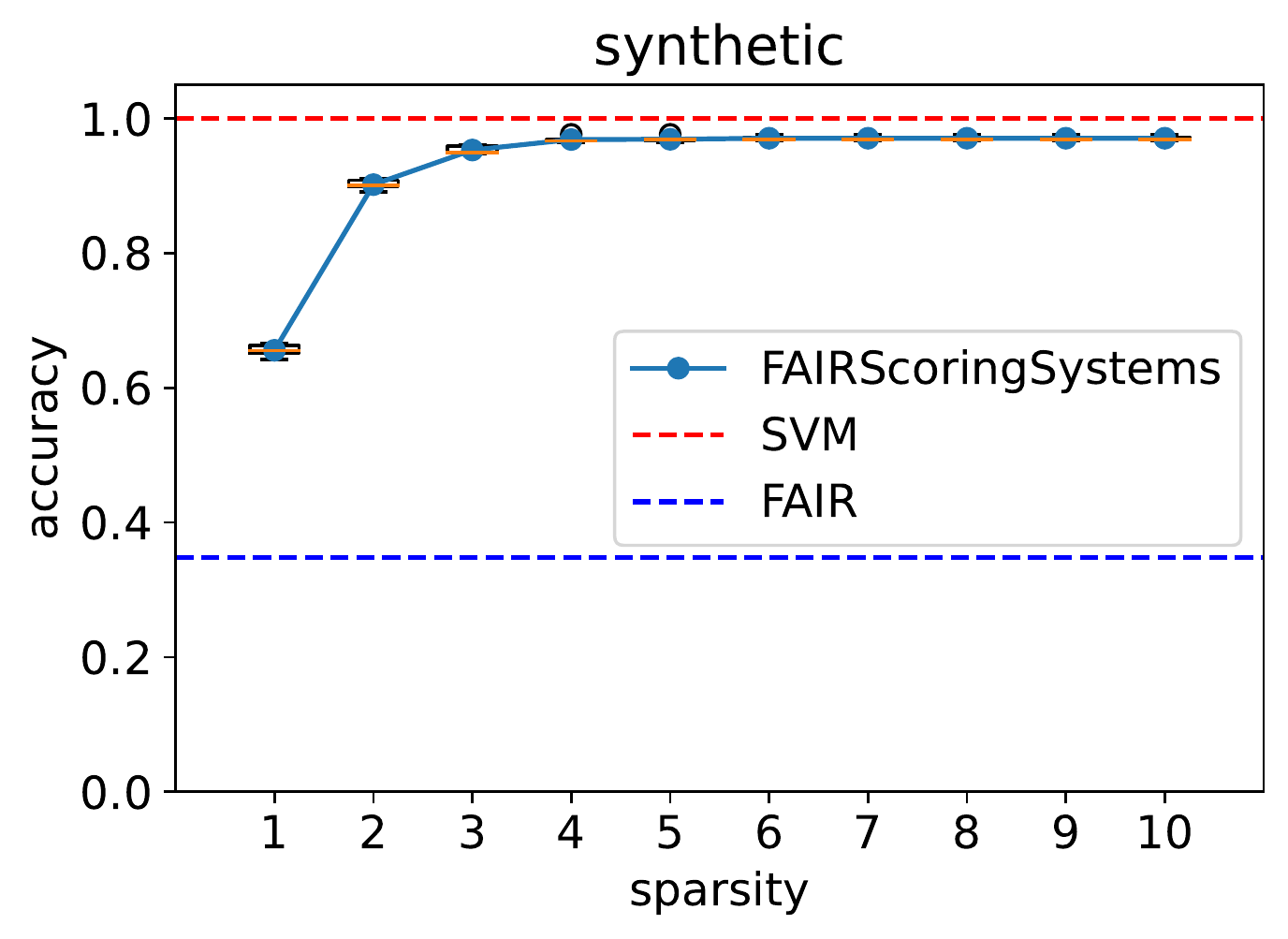}}%
    \hfill
    \subcaptionbox{wine dataset}{\includegraphics[width=0.31\textwidth]{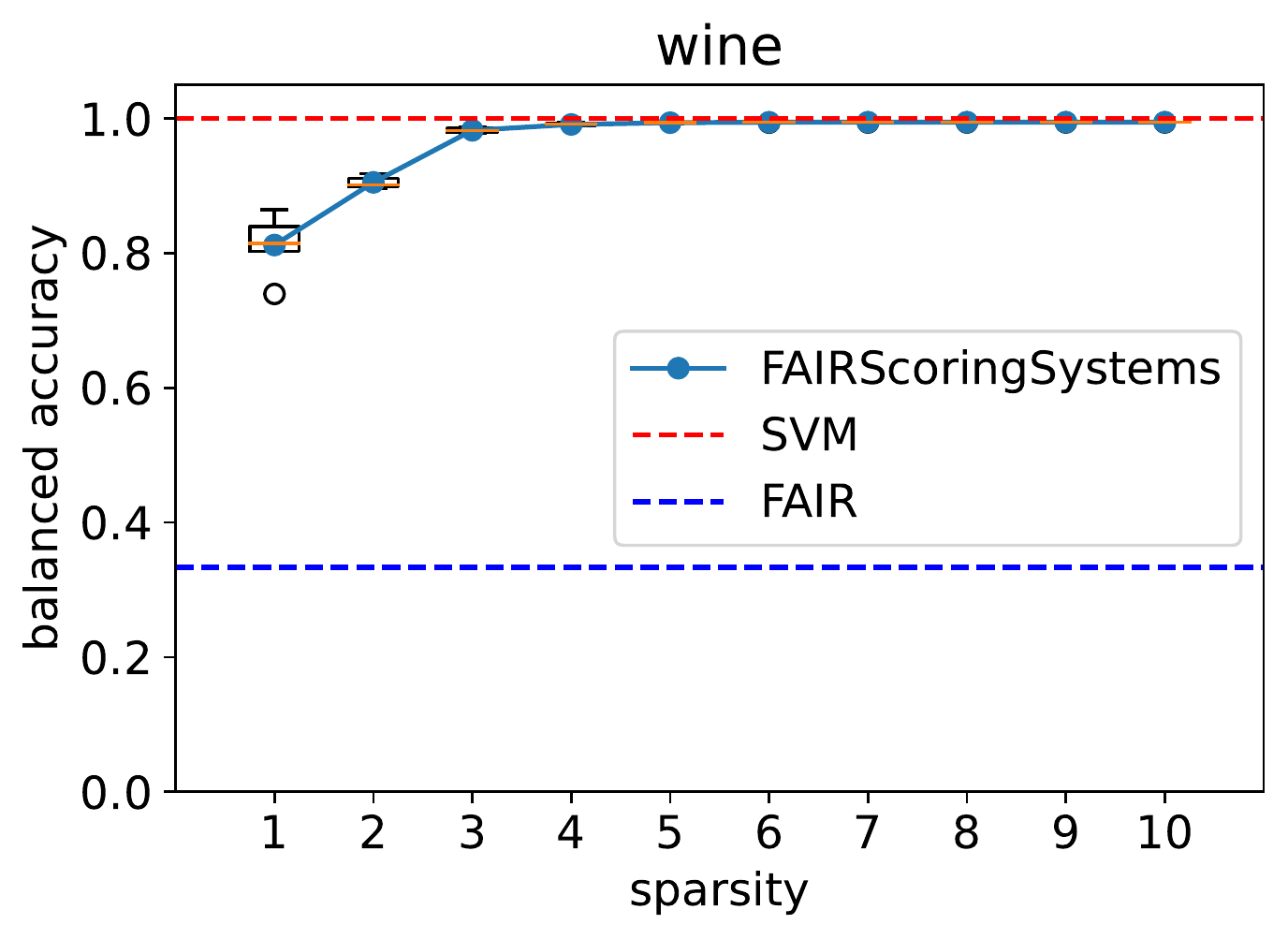}}%
    \hfill
    \subcaptionbox{customer dataset}{\includegraphics[width=0.31\textwidth]{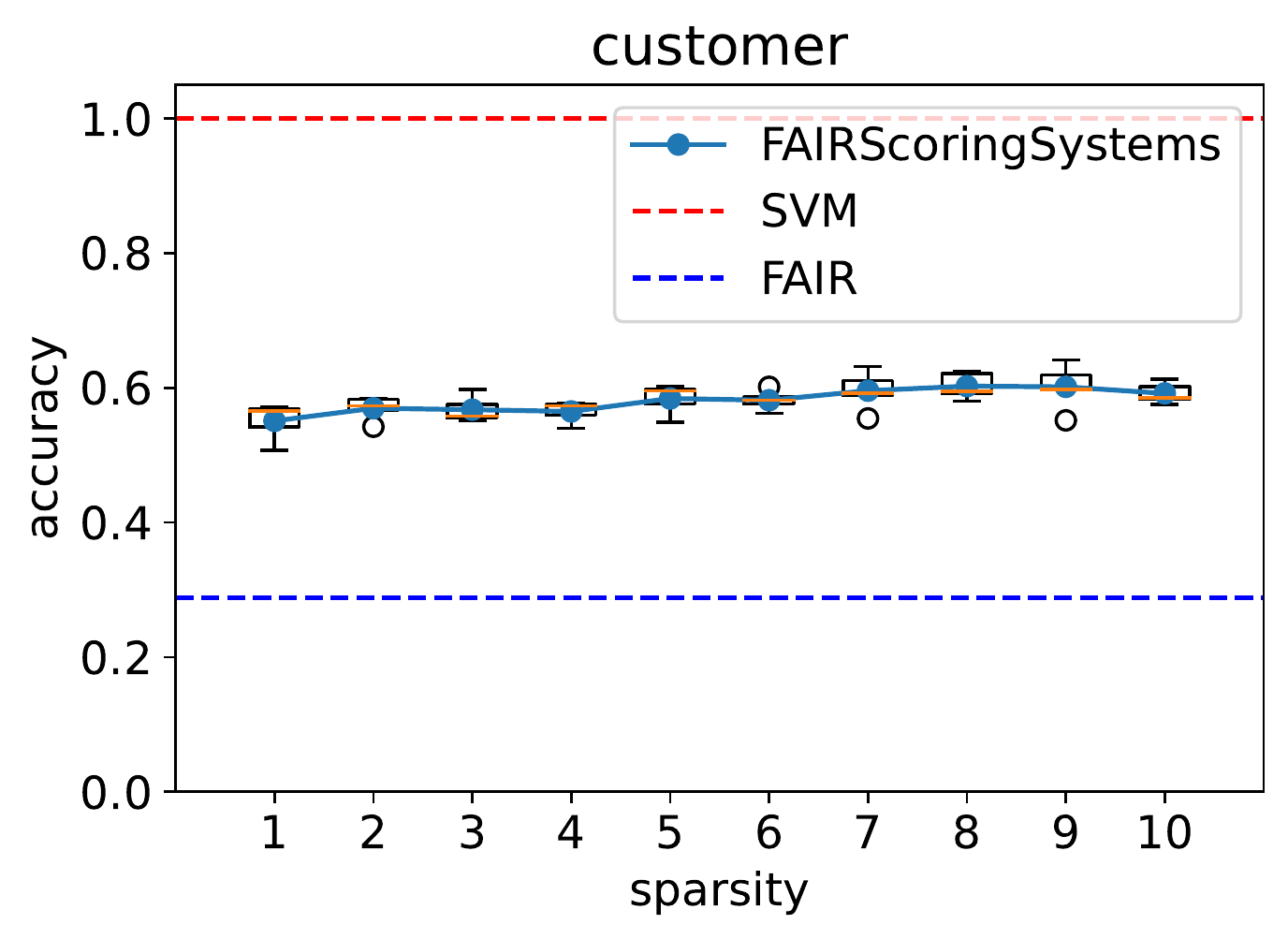}}%
    \hfill
    \caption{Accuracy (train set) as a function of the sparsity constraint (maximum number of rules in each scoring system).}    
    \label{Fig:acc_train}
\end{figure*}

\begin{figure*}[!htb]
    \centering
    \subcaptionbox{synthetic dataset}{\includegraphics[width=0.31\textwidth]{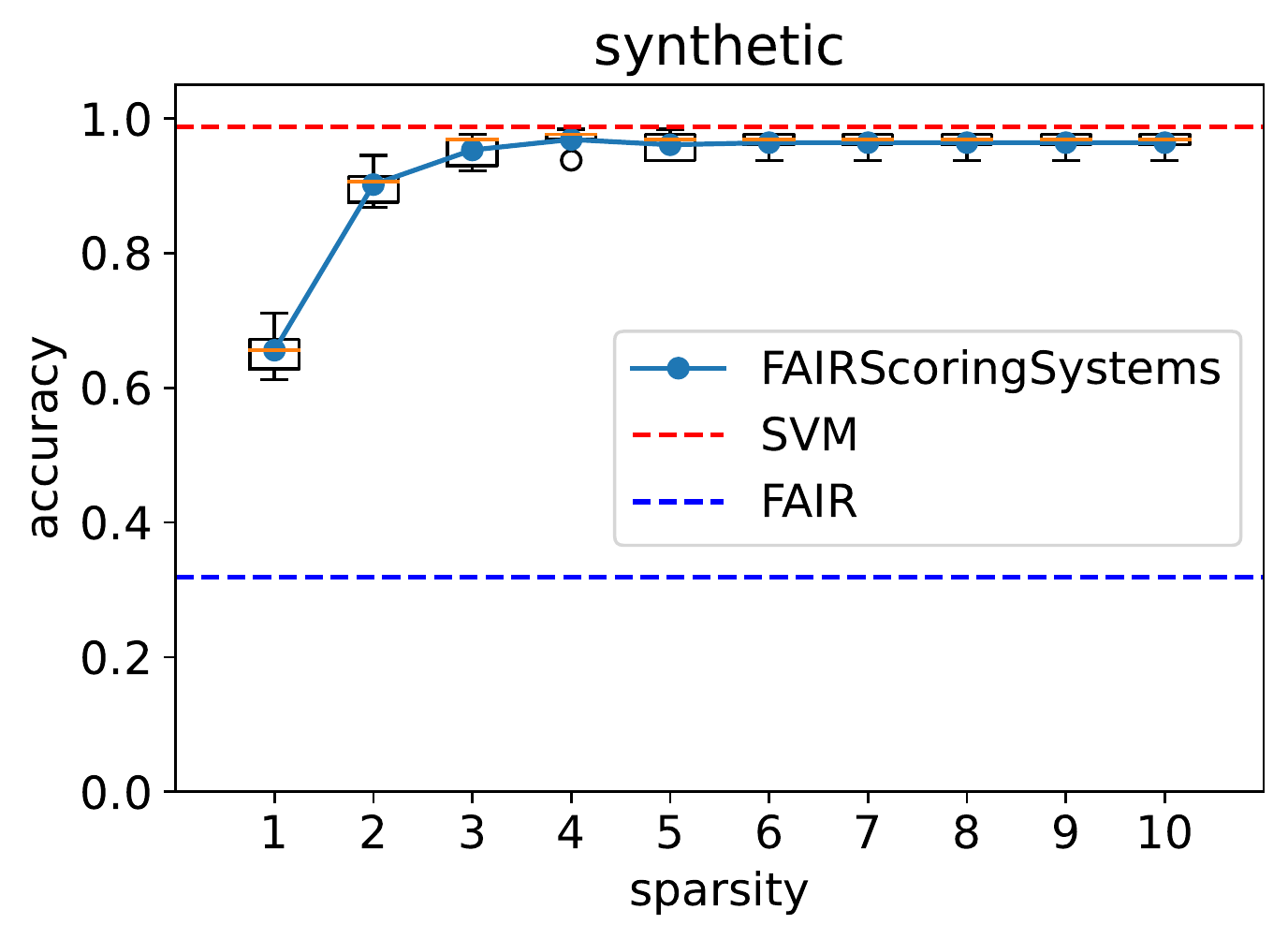}}%
    \hfill
    \subcaptionbox{wine dataset}{\includegraphics[width=0.31\textwidth]{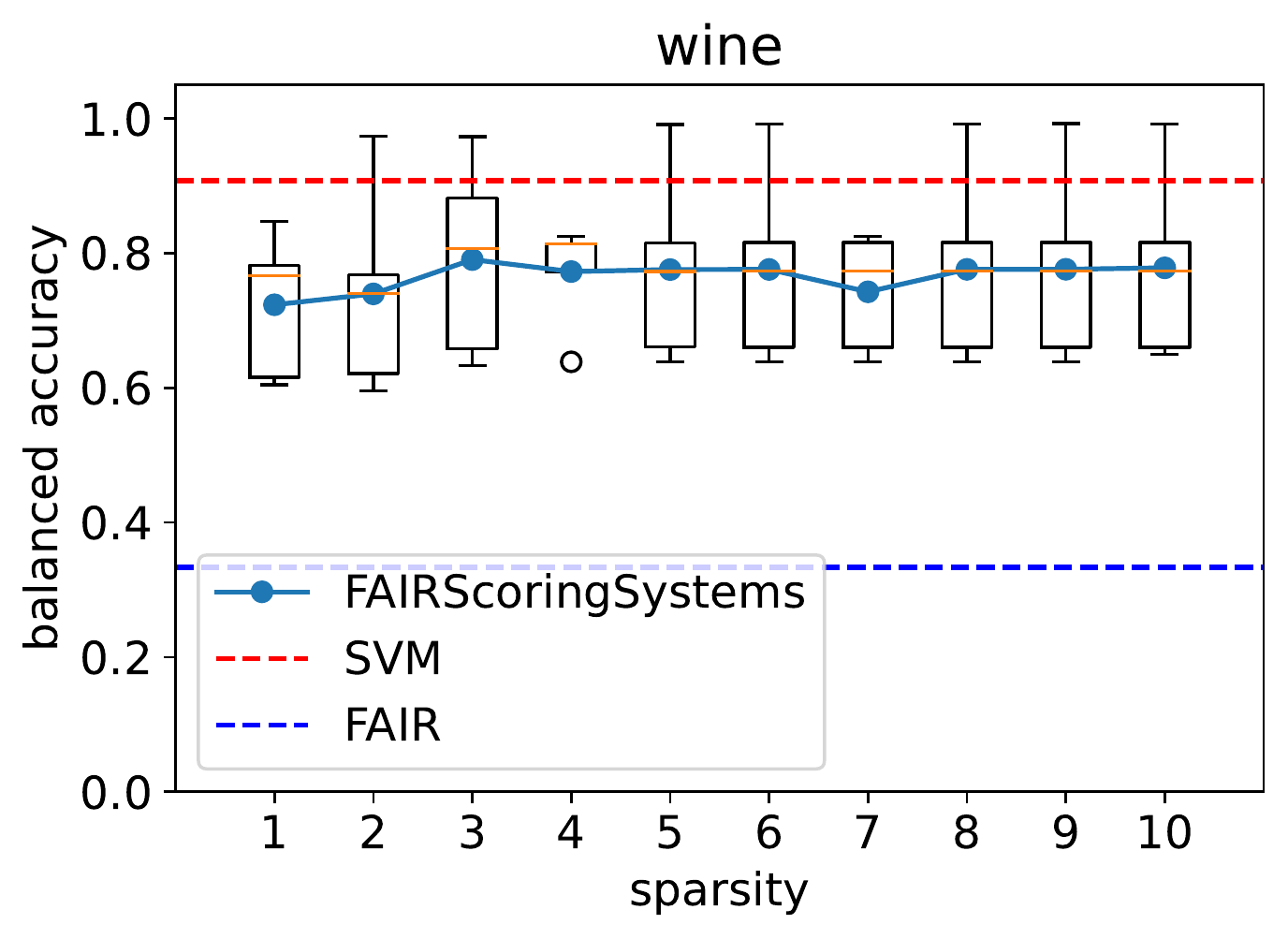}}%
    \hfill
    \subcaptionbox{customer dataset}{\includegraphics[width=0.31\textwidth]{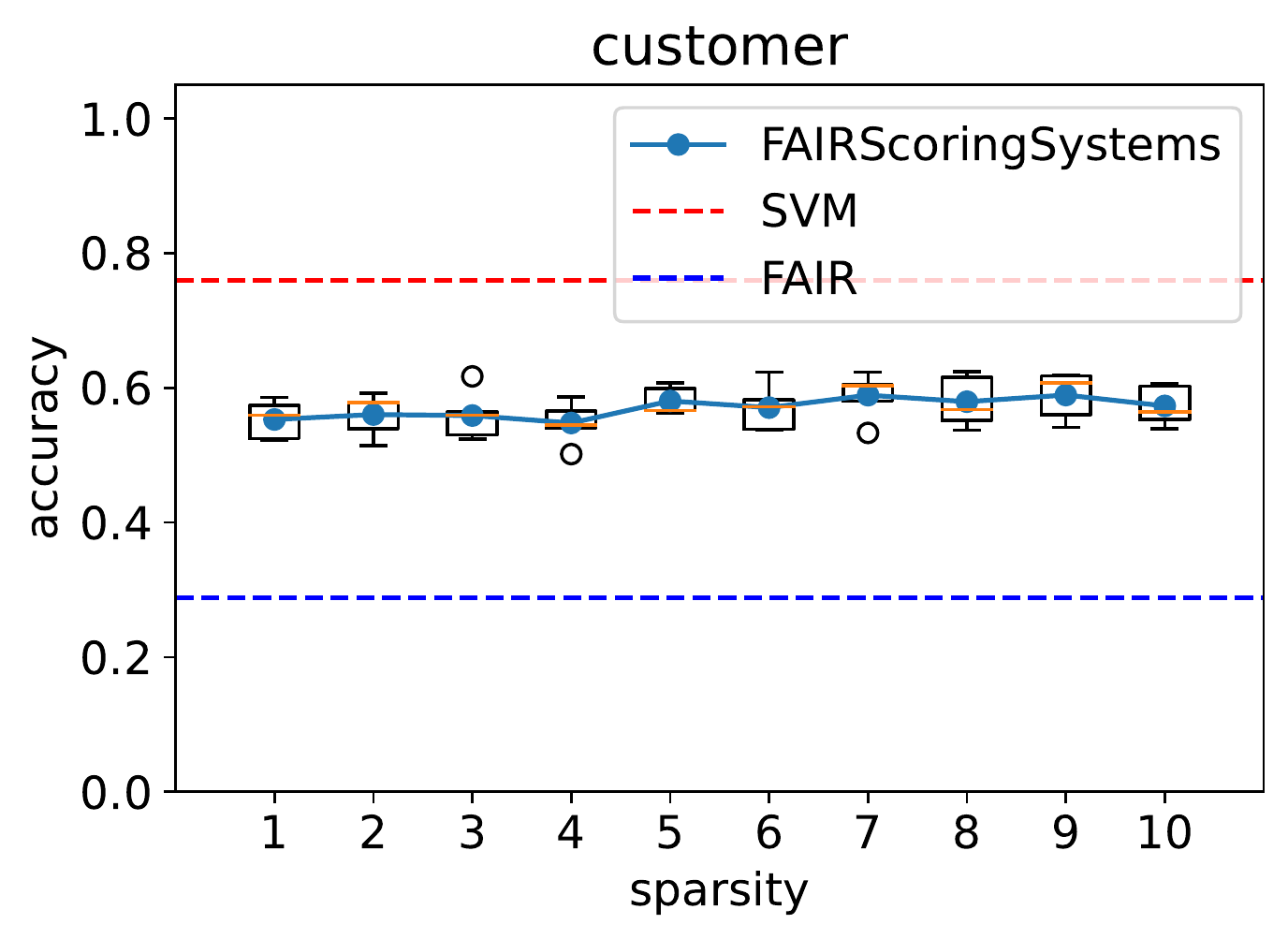}}%
    \hfill
    \caption{Accuracy (test set) as a function of the sparsity constraint (maximum number of rules in each scoring system).}
    \label{Fig:acc_test}
\end{figure*}

We first evaluate the accuracy of the produced multi-class scoring systems as a function of the sparsity constraint (\emph{i.e.}, maximum number of lines for each individual scoring system).
Figure~\ref{Fig:acc_train} (respectively, \ref{Fig:acc_test}) displays the performances on the training set (respectively, test set) for the three datasets with sparsity limits ranging from 1 to 10.

\begin{figure*}[!htb]
    \centering
    \subcaptionbox{synthetic dataset}{\includegraphics[width=0.31\textwidth]{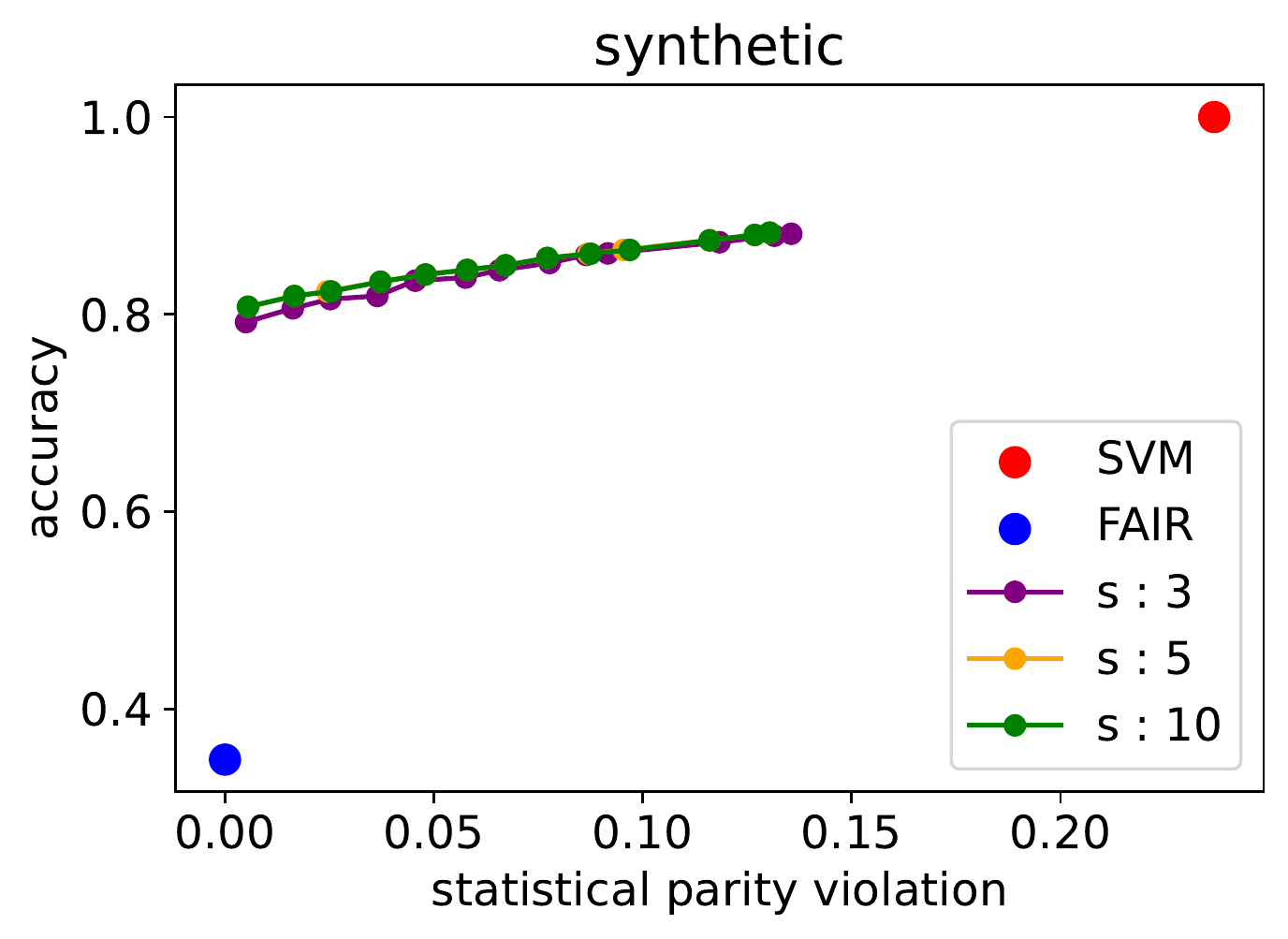}}%
    \hfill
    \subcaptionbox{wine dataset}{\includegraphics[width=0.31\textwidth]{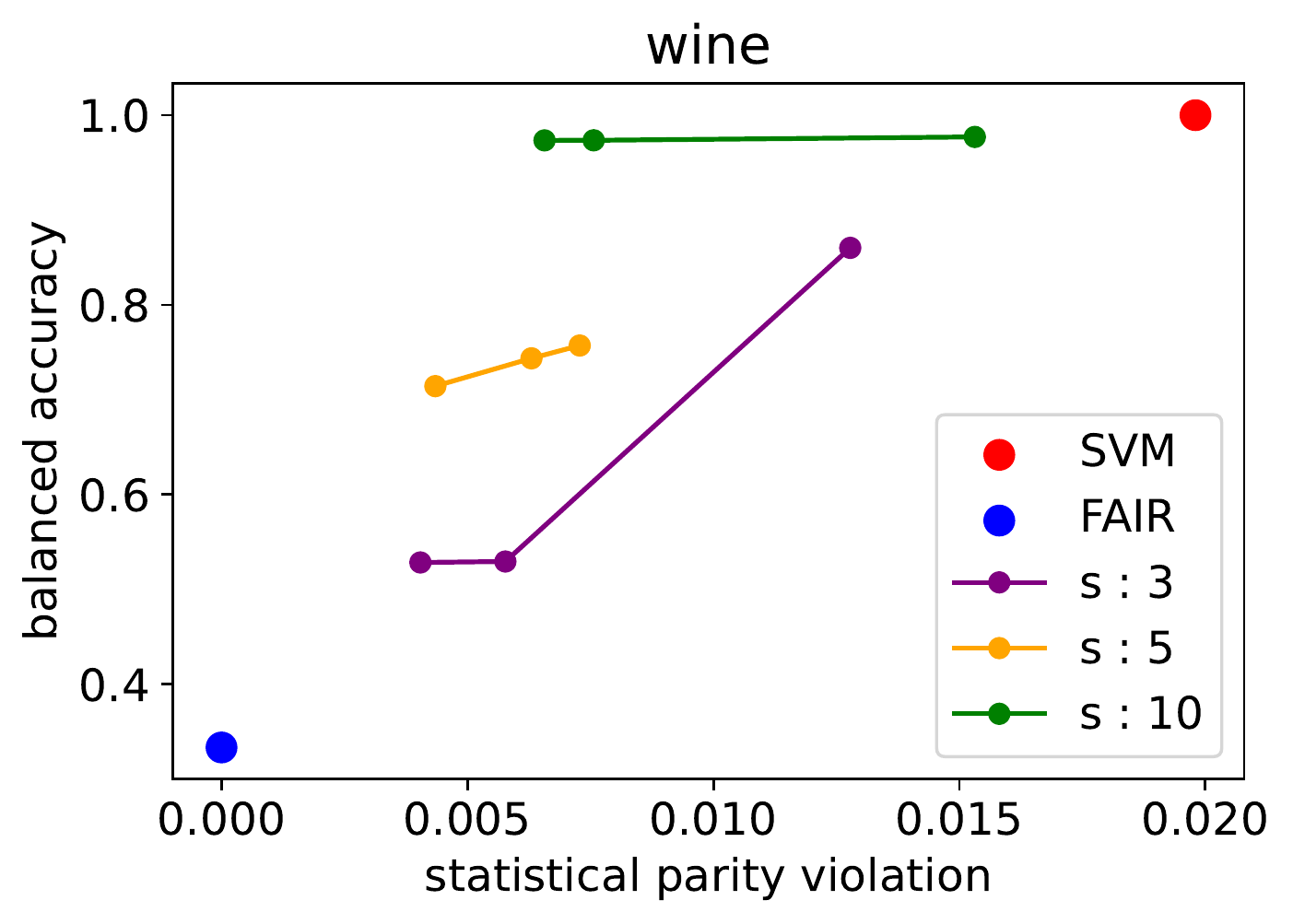}}%
    \hfill
    \subcaptionbox{customer dataset}{\includegraphics[width=0.31\textwidth]{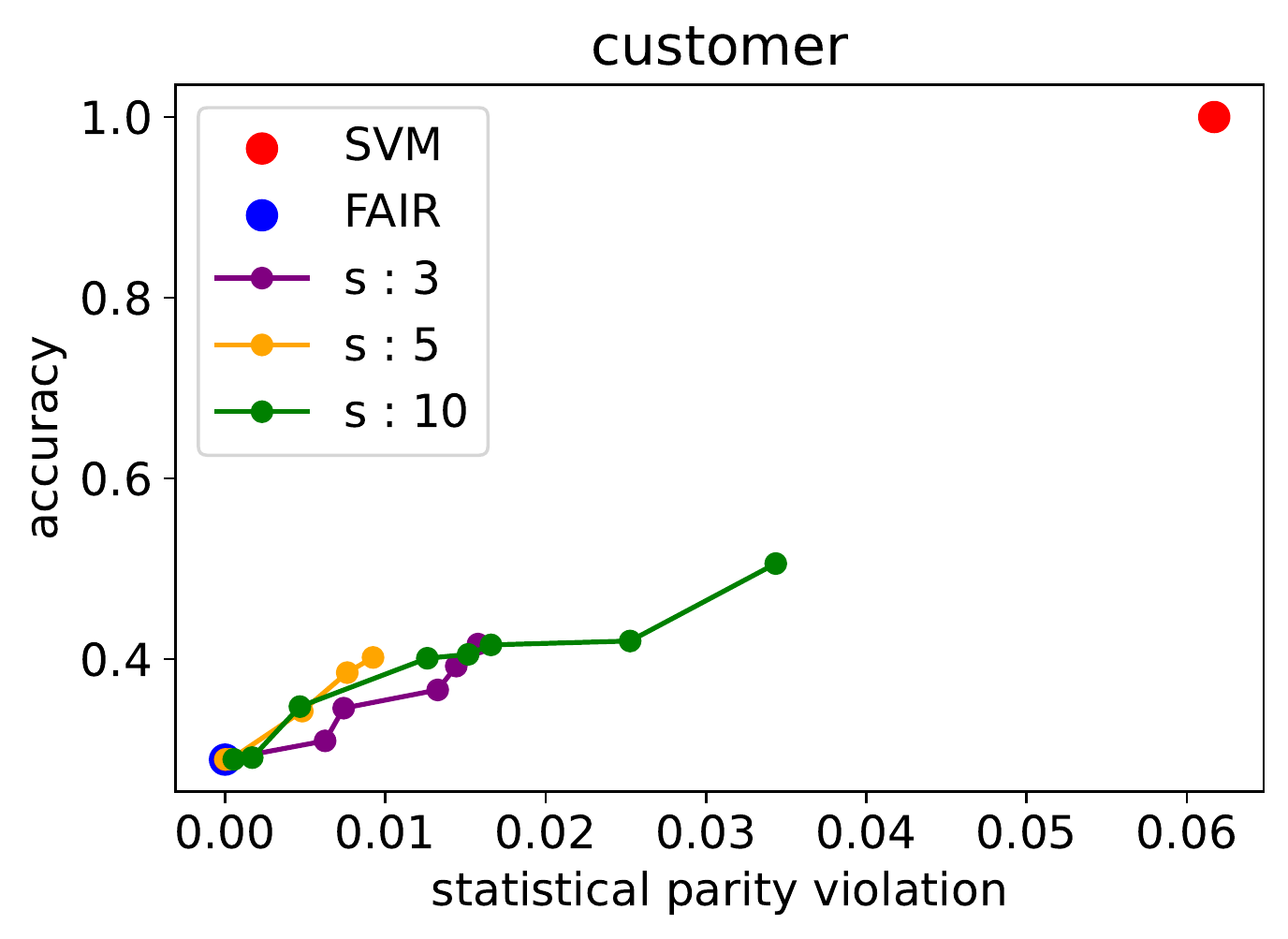}}%
    \hfill
    \caption{Accuracy (train set) as a function of the unfairness value (statistical parity), for different sparsity constraints (3, 5, 10).}    
    \label{Fig:sp_train}
\end{figure*}

\begin{figure*}[!htb]
    \centering
    \subcaptionbox{synthetic dataset}{\includegraphics[width=0.31\textwidth]{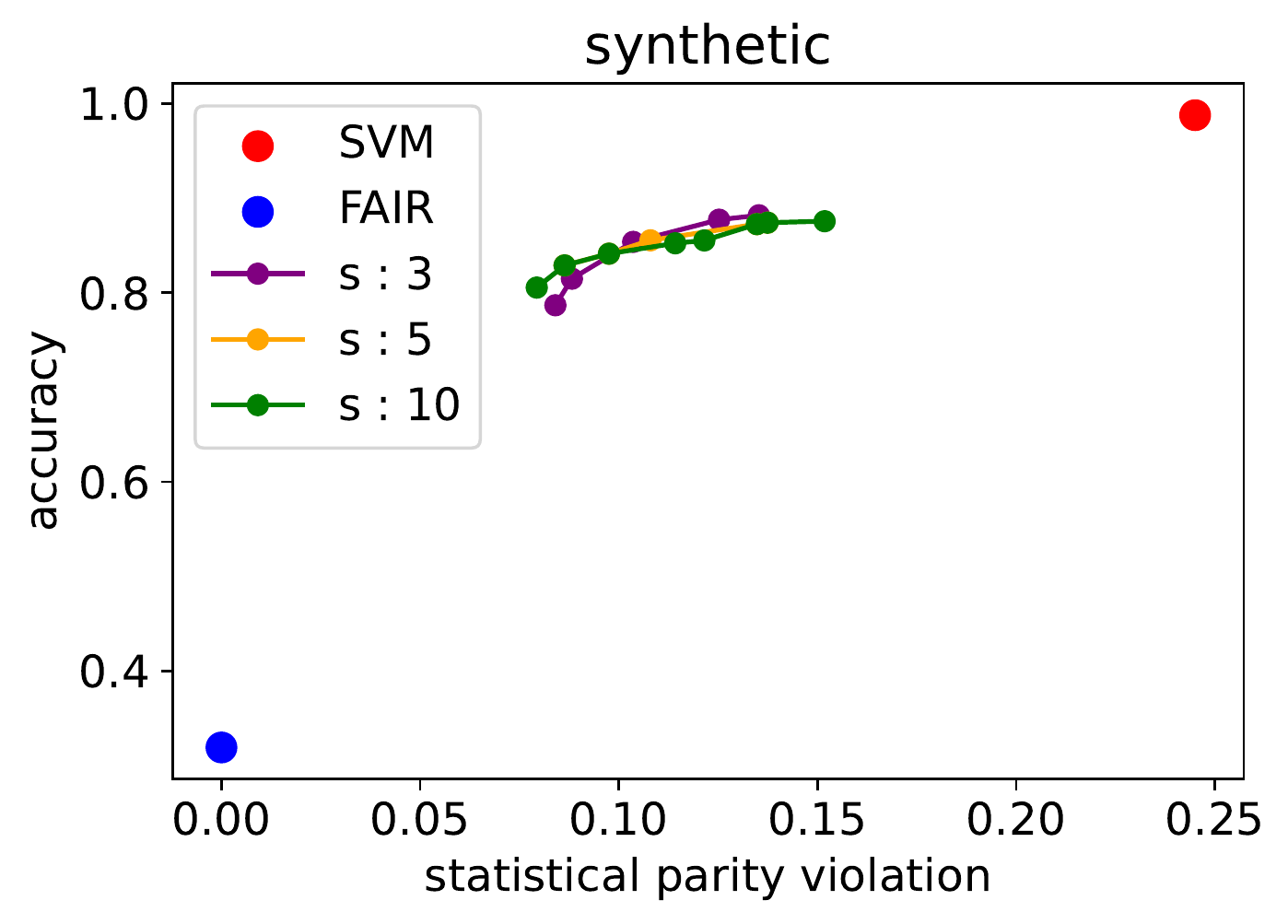}}%
    \hfill
    \subcaptionbox{wine dataset}{\includegraphics[width=0.31\textwidth]{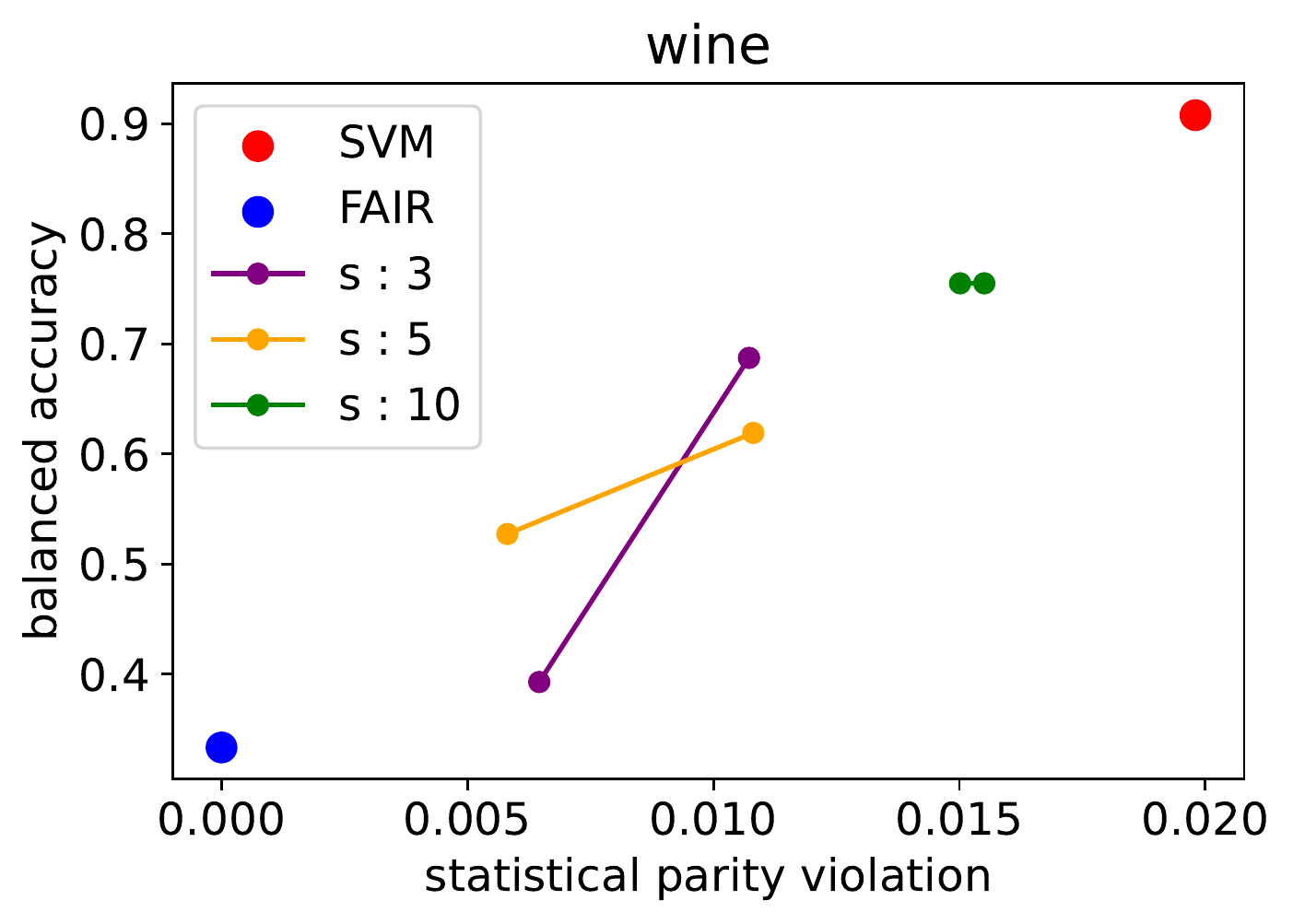}}%
    \hfill
    \subcaptionbox{customer dataset}{\includegraphics[width=0.31\textwidth]{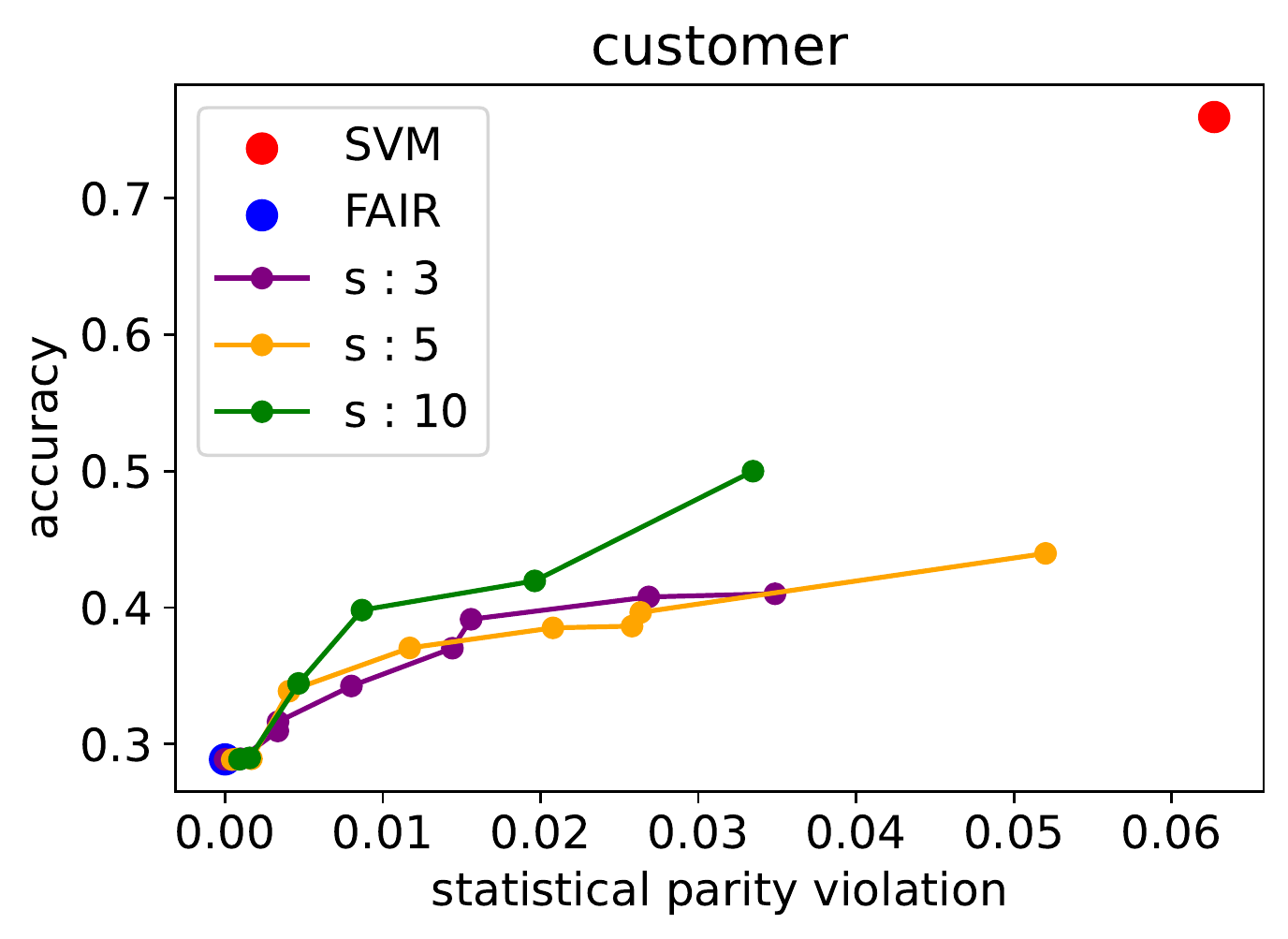}}%
    \hfill
    \caption{Accuracy (test set) as a function of the unfairness value (statistical parity), for different sparsity constraints (3, 5, 10).} 
    \label{Fig:sp_test}
\end{figure*}

Figures~\ref{Fig:acc_train} and \ref{Fig:acc_test} show that while very tight sparsity constraints can have a considerable impact on the models' accuracy, the use of \emph{reasonably} sparse interpretable models results in good performances.
For the synthetic dataset, all experiments produced certifiably optimal scoring systems, while for the wine dataset, optimality gaps always lied below 0.1\%.

\subsection{Fairness/Accuracy Trade-offs}
We then evaluate our framework on the multi-class fairness metrics introduced in section~\ref{section:contributions_metrics}, for a wide range of unfairness tolerances (up to 20 values of $\epsilon$ ranging non linearly between 1\% and the \texttt{SVM} unfairness, with a higher density near 1\%), and different sparsity constraints (3, 5 and 10).
Due to the limited space available, we only report results for the multi-class statistical parity metric.
Nonetheless, the results obtained for the other metrics summarized in Table~\ref{tab:proposed-multi-fairness-metrics} show similar trends and are available on our repository.

Figure~\ref{Fig:sp_train} (respectively, Figure~\ref{Fig:sp_test}) provides the Pareto frontiers between accuracy 
and fairness violation for the multi-class statistical parity metric on the training set (respectively, test set). 
The results show that FAIRScoringSystems is able to produce interpretable models exhibiting interesting trade-offs between accuracy and fairness. 
For reasonable sparsity constraints, the generated models allow for substantial fairness violation reductions (compared to \texttt{SVM}) while still exhibiting good classification performances. 
Given the accuracy and fairness values, along with the interpretable models themselves, a domain-expert may then pick the most relevant model among the produced frontier. 
\section{Conclusion}\label{section:conclusion}
In this paper, we have proposed FAIRScoringSystems, a Mixed Integer Linear Programming framework producing optimal scoring systems, which are inherently interpretable models, under fairness and sparsity constraints for multi-class classification. 
To the best of our knowledge, this is the first work tackling these different requirements altogether.
Our experimental evaluation demonstrates that FAIRScoringSystems is able to generate interesting trade-offs between accuracy, fairness and sparsity on both synthetic and real-world multi-class classification datasets of various shapes.

Future work includes improving our framework's scalability, although it is already able to learn well-performing models for real-size datasets.
While reaching and proving optimality for difficult datasets (\emph{i.e.}, non-linearly separable, with high numbers of samples and features) is computationally challenging, our method can still be used to produce well-performing models.
Using off-the-shelf solvers, our framework works in an \emph{any-time} fashion: even if the learning process is stopped before proving optimality, a valid solution (\emph{i.e.,} a multi-class scoring system satisfying sparsity and fairness constraints) can be returned. This implies that it can be used to learn fair and accurate interpretable models given limited time budgets. 
The produced models additionally come with a quality guarantee provided by optimality gaps, which upper-bound the distance to the actual optimal solution.

\bibliographystyle{IEEEtran}
\bibliography{IEEEabrv,bibliography.bib}

\end{document}